\documentclass[letterpaper, 10 pt, conference]{ieeeconf}  

\IEEEoverridecommandlockouts                              

\usepackage{graphics} 
\usepackage{amsmath,mathtools,amsfonts,amssymb} 
\usepackage{tcolorbox}
\usepackage{tabularx}
\usepackage{booktabs}
\usepackage{BOONDOX-cal}
\usepackage{multirow}
\usepackage{cite} 

\usepackage{xcolor} 

\usepackage{gensymb} 

\usepackage{algorithm2e}
\RestyleAlgo{ruled}

\usepackage{todonotes}
\usepackage[caption=false,font=footnotesize]{subfig}

\usepackage[hidelinks]{hyperref}

\title{\LARGE \bf	 
	 Enhancing motion trajectory segmentation of rigid bodies \\ using a novel screw-based trajectory-shape representation
	}

\author{Arno Verduyn$^{1,2}$, Maxim Vochten$^{1}$, Joris De Schutter$^{1}$
\thanks{This result is part of a project that has received funding from the European Research Council (ERC) under the European Union's Horizon 2020 research and innovation programme (Grant agreement No. 788298).}
\thanks{$^{1}$Department of Mechanical Engineering and Flanders Make at KU Leuven, 3001 Leuven, Belgium.}%
\thanks{$^{2}$Corresponding author (arno.verduyn@kuleuven.be)}
}

\begin{document}


\maketitle
\thispagestyle{empty}
\pagestyle{empty}

\begin{abstract}

Trajectory segmentation refers to dividing a trajectory into meaningful consecutive sub-trajectories.
This paper focuses on trajectory segmentation for 3D rigid-body motions. Most segmentation approaches in the literature represent the body's trajectory as a point trajectory, considering only its translation and neglecting its rotation. 
We propose a novel trajectory representation for rigid-body motions that incorporates both translation and rotation, and additionally exhibits several invariant properties. This representation consists of a geometric progress rate and a third-order trajectory-shape descriptor.
Concepts from screw theory were used to make this representation time-invariant and also invariant to the choice of body reference point. 
This new representation is validated for a self-supervised segmentation approach, both in simulation and using real recordings of human-demonstrated pouring motions.
The results show a more robust detection of consecutive sub-motions with distinct features and a more consistent segmentation 
compared to conventional representations.  
We believe that other existing segmentation methods may benefit from using this trajectory representation to improve their invariance. 

\end{abstract}


\section{INTRODUCTION}
Trajectory segmentation aims to divide a trajectory into consecutive sub-trajectories that have a specific meaning for the considered application. It is a valuable tool to reduce the dimensionality of trajectories, thereby improving the efficiency and reliability of trajectory recognition and learning algorithms. 
Trajectory segmentation has found applications in various research fields, including monitoring urban traffic \cite{DODGE2009}, tracking changes in animal behavior \cite{BehaviouralScience}, analyzing robot-assisted minimally invasive surgical procedures \cite{UnsupervisedSegmentationForRobotLearning}, and learning motion primitives for robots from human-demonstrated object manipulation tasks \cite{HierarchicalSegmentationOfManipulationTasks,AutonomousFrameworkForSegmentingManipulationTasks,UnsupervisedTrajectorySegmentation,OnlineSegmentationAndClustering}.



In the literature, three types of segmentation approaches can be distinguished: supervised segmentation, unsupervised segmentation, and self-supervised segmentation. 

\textit{Supervised segmentation} algorithms rely on prior expert knowledge, such as a predefined library of template segments, such as in \cite{templatelibrary}, or a set of predefined segmentation criteria. The approaches in \cite{DODGE2009,SegmentingBasedOnSpatioTemporalCriteria} segment trajectories based on criteria that assess the homogeneity of movement profiles, including the location, speed and heading of the object. 

\textit{Unsupervised segmentation} algorithms do not rely on prior expert knowledge. These approaches are typically event-based, searching for discrete segmentation points along the trajectory.
The approach in \cite{AutonomousFrameworkForSegmentingManipulationTasks} identifies segmentation points by fitting a Gaussian Mixture Model to the trajectory data and identifying regions of overlap between the confidence ellipsoids of consecutive Gaussians. Another approach involves extracting segmentation points from local extrema in curvature and torsion along the trajectory, such as in \cite{UnsupervisedTrajectorySegmentation}. 

\textit{Self-supervised segmentation} can be viewed as a supervised approach in which the segmentation criteria are learned from the data rather than being predetermined by expert knowledge. The segmentation approach in \cite{FeatureExtractionBasedSegmentation} detects transitions in movement profiles and labels them as `features'. Similar features are then clustered together, and subsequently, the trajectory is segmented into a sequence of cluster labels.  
The approach in \cite{OnlineSegmentationAndClustering} initially segments trajectories in an unsupervised way. It then learns a library of motion primitives from the segmented data, which is subsequently used to improve the initial segmentation accuracy. 




The literature on trajectory segmentation has two main shortcomings. The first shortcoming is that the object is typically approximated as a moving point, considering only the translational motion while neglecting the rotational motion. Hence, not all trajectory information is taken into account. 

The second shortcoming is that supervised and self-supervised segmentation approaches based on template matching typically are dependent on the motion profile of the trajectory and the choice of references. These references include both the coordinate frame in which the trajectory coordinates are expressed and the reference point on the body being tracked. This dependency limits the approach's capability to generalize across different setups. 

The main objective of this work is to enhance template-based approaches in supervised and self-supervised trajectory segmentation by incorporating invariance with respect to time, coordinate frame, and body reference point.

Our approach consists of first reparameterizing the trajectory to achieve time-invariance, i.e. invariance to changes in motion profile. This is achieved by defining a novel geometric progress rate using Screw Theory, inspired by \cite{Joris2010}. The progress rate combines both rotation and translation, while also being invariant to the chosen body reference point. The defined progress rate is regularized to better cope with pure translations compared to \cite{Joris2010}. After reparameterization, the trajectory is represented using a novel trajectory-shape descriptor based on first-order kinematics of the trajectory. 
Finally, invariance to changes in the coordinate frame is achieved by performing a spatial alignment before matching the trajectory descriptor with template segments in the library. The latter template segments are also represented using the same trajectory-shape descriptor.

The contribution of this work is (1) the introduction of a novel \textit{screw-based geometric progress rate} for rigid-body trajectories, and (2) a novel \textit{trajectory-shape descriptor}. These concepts were applied in a self-supervised segmentation approach using simulations and real recordings of human-demonstrated pouring motions. The results show that more consistent segmentation results can be achieved by using an invariant trajectory descriptor compared to conventional descriptors that are not invariant to changes in execution speed, coordinate frame, and reference point.  

The outline of the paper is as follows. Section \ref{sec:preliminaries} reviews essential background on rigid-body trajectories and screw theory. Section \ref{sec:approach} introduces the novel screw-based geometric progress rate and trajectory-shape descriptor for rigid-body trajectories. Section \ref{sec:segmentation} explains the implementation of these novel concepts in a self-supervised segmentation approach.  Section \ref{sec:experiments} explains the validation of the segmentation approach using simulations and real recordings of human-demonstrated pouring motions. Section \ref{sec:discussion} concludes with a discussion and suggests future work.

\section{PRELIMINARIES}
\label{sec:preliminaries}



The displacement of a rigid body in 3D space is commonly represented by attaching a body frame to the rigid body and expressing the position $\boldsymbol{p}$ and orientation $R$ of this frame with respect to a fixed world frame. The position coordinates $\boldsymbol{p}$ represent the relative position of the origin of the body frame with respect to the origin of the world frame. The rotation matrix $R$, which is part of the Special Orthogonal group SO(3), represents the relative orientation of the body frame with respect to the world frame\footnote{Note that other possibilities exist for representing the orientation such as the quaternion or the axis-angle representation.}. The position $\boldsymbol{p}$ and orientation $R$ of the rigid body can be combined into the homogeneous transformation matrix $\footnotesize T=	\begin{bmatrix} R & \boldsymbol{p} \\ \boldsymbol{0} & 1 \end{bmatrix}$, which is part of the Special Euclidean group $SE(3)$. The homogeneous transformation matrix as a function of time $T(t)$ represents a temporal rigid-body trajectory.

The first-order kinematics of the rigid-body trajectory $T(t)$ are commonly represented by a 6D twist $\boldsymbol{\mathcal{t}}$ consisting of two 3D vectors: a rotational velocity vector $\boldsymbol{\omega}$ and a linear velocity vector $\boldsymbol{v}$. Based on the coordinate frame in which these velocities are expressed and based on the reference point for the linear velocity $\boldsymbol{v}$, three twists are commonly defined in the literature \cite{lynch2017modern,HermanCursus}, i.e. the \textit{pose twist}, \textit{spatial twist}, and \textit{body twist}. A twist is considered left- or right-invariant when it is invariant to changes of the world or body frame, respectively. The \textit{body twist} is left-invariant, the \textit{spatial twist} is right-invariant, and the \textit{pose twist} is neither left- nor right-invariant.

In rigid-body kinematics, the Mozzi-Chasles' theorem \cite{Chasles,Ceccarelli} states that a rigid-body velocity can always be represented as a rotation about an axis in space and a translation parallel to this axis, referred to as the \textit{Instantaneous Screw Axis} (ISA). The direction of the ISA is uniquely defined by the direction of the rotational velocity $\boldsymbol{\omega}$, while the location of the ISA is defined by any point on the ISA. A unique choice is to take the point on the ISA that lies closest to the reference point for the linear velocity $\boldsymbol{v}$. This point's position vector $\boldsymbol{p}_\perp$  as well as the translational velocity parallel to the ISA, $\boldsymbol{\nu}$, are calculated from the twist components $\boldsymbol{\omega}$ and $\boldsymbol{v}$:
\begin{equation}
\label{eq:nu}
\boldsymbol{p}_\perp = \frac{ \boldsymbol{\omega}\times\boldsymbol{v}}{\Vert \boldsymbol{\omega}\Vert^2} ~~~~ \text{and} ~~~~ \boldsymbol{\nu} = \boldsymbol{v} + \boldsymbol{\omega} \times \boldsymbol{p}_\perp.
\end{equation}
When the \textit{pose twist} is chosen, the coordinates of $\boldsymbol{p}_\perp$ and $\boldsymbol{\nu}$ are in the world frame, while $\boldsymbol{p}_\perp$ is the position vector from the body frame's origin to the closest point on the ISA.


The magnitudes of the rotational and translational velocities $\Vert \boldsymbol{\omega}\Vert$ and $\Vert \boldsymbol{\nu}\Vert$ are closely related to the `SE(3) invariants' in \cite{SE3metrics} and it has been shown that they are both left- and right-invariant, also referred to as \textit{bi-invariant}. 



\section{SCREW-BASED TRAJECTORY REPRESENTATION}
\label{sec:approach}

This section introduces a novel screw-based geometric progress rate for defining the progress over a trajectory, independent of time. Next, a new invariant trajectory-shape descriptor is introduced for rigid-body trajectories, which is both time-invariant and invariant to the choice of the reference point on the body. Invariance to changes in coordinate frame is obtained by proposing a spatial alignment algorithm.




\subsection{Screw-based geometric progress rate}
\label{sec:progress_rate}

\begin{figure}[t]
	\centering
	\vspace{5pt}
	\includegraphics[width=0.45\linewidth]{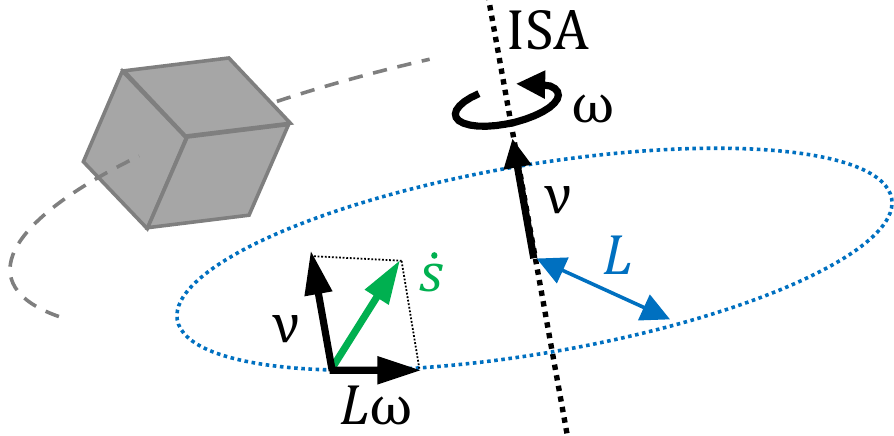}%
	\caption{Interpretation of the proposed progress rate $\dot{s}$ as the linear velocity of a point on the moving body at a distance $L$ from the ISA. 
	}
	\label{fig:definition_progress}
\end{figure}

To obtain a time-invariant representation of the trajectory, a geometric progress rate has to be defined first. We propose to combine the magnitude of the rotational velocity $\Vert \boldsymbol{\omega} \Vert$ and the translational velocity $\Vert\boldsymbol{\nu}\Vert$ parallel to the ISA:
\begin{equation}
	\label{eq:progress_rate}
	\dot{s} = \sqrt{L^2\Vert \boldsymbol{\omega} \Vert^2 + \Vert \boldsymbol{\nu} \Vert^2} ~,
\end{equation}
where $L$ is a weighting factor with units [m]. The progress rate $\dot{s}$ [m/s] can be interpreted as the magnitude of the linear velocity of any point on the body, at a distance $L$ from the ISA (see Fig.~\ref{fig:definition_progress}). The progress parameter $s$ [m] is then found as the integral of the progress rate over time, and can be considered as a scalar value signifying the geometric progress in translation and rotation over the trajectory.

\begin{figure}[t]
	\centering
	\includegraphics[width=0.4\linewidth]{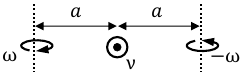}%
	\caption{Illustration of a pure couple of rotations $\boldsymbol{\omega}$ generating a pure translation $\boldsymbol{\nu}$ out of the plane. The two rotation axes, distanced by two times $a$, are depicted with dotted lines.}
	\label{fig:rotation_couple}
	\vspace{-5pt}
\end{figure}

Since $\Vert \boldsymbol{\omega}\Vert$ and $\Vert \boldsymbol{\nu}\Vert$ are bi-invariant properties of the trajectory, as mentioned in Section \ref{sec:preliminaries}, the resulting geometric progress rate $\dot{s}$ is also bi-invariant. However, it is important to clarify that the proposed progress rate does not comply with the conditions for being a bi-invariant metric on $se(3)$ \cite{SE3metricsPark}. For example, consider a serial kinematic chain of two twist motions $\boldsymbol{\mathcal{t}}_1$ and $\boldsymbol{\mathcal{t}}_2$ generating a resulting motion $\boldsymbol{\mathcal{t}}_{1+2}$. Then, the triangle inequality $\dot{s}_{1+2} \leq \dot{s}_1 + \dot{s}_2$ (a necessary condition for being a metric) does not hold for all cases on $se(3)$. 
Consider the example of a pure translation $\dot{s}_{1+2} = \Vert \boldsymbol{\nu} \Vert$. This translation can be generated by a couple of rotations with magnitude $\Vert \boldsymbol{\omega} \Vert = \frac{\Vert \boldsymbol{\nu} \Vert}{2a}$, where $a$ is half the distance between the two rotation axes, as shown in Fig.~\ref{fig:rotation_couple}.

\noindent Since the sum of the progress rates of these rotations equals:   
\begin{equation}
	\label{eq:example_couple}
	\dot{s}_1 + \dot{s}_2 = 2L\Vert \boldsymbol{\omega}\Vert = \frac{L}{a}\Vert \boldsymbol{\nu} \Vert,
\end{equation}
it can be concluded that $\dot{s}_1 + \dot{s}_2 < \dot{s}_{1+2}$, if $L < a$.  Since $L$ is a predefined weighting factor and $a$ can take any value in $\mathbb{R}^+_0$ 
(because $\boldsymbol{\omega}$ can always be decreased by increasing $a$ while still resulting in the same $\boldsymbol{\mathcal{\nu}}$),
there is no choice for $L$ for which the triangle inequality holds for each value of $a$. 


\subsection{Regularized geometric progress rate}
\label{sec:regul_progress}
\begin{figure}[t]
	\centering
	\vspace{5pt}
	\includegraphics[width=0.35\linewidth]{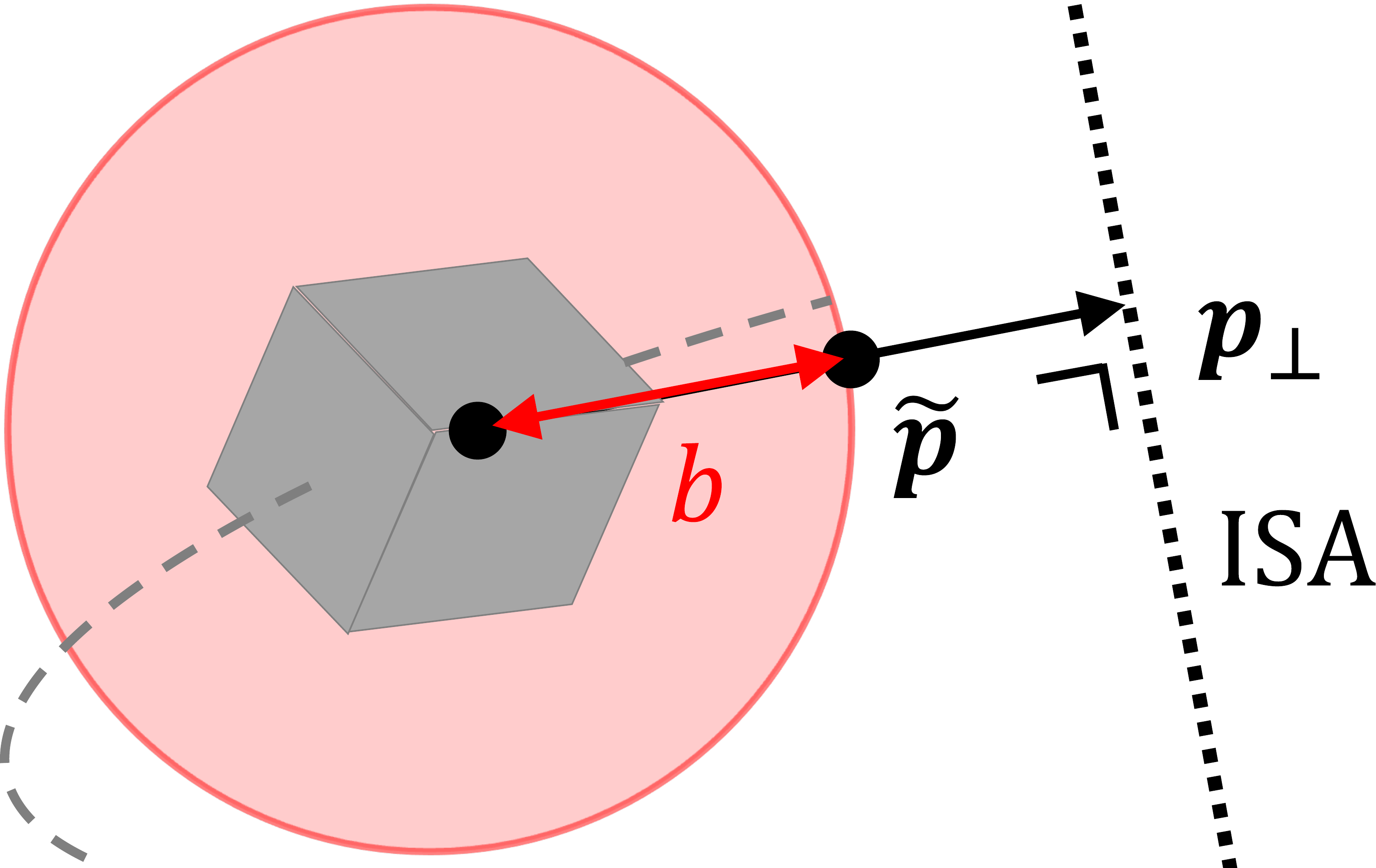}%
	\caption{Point $\boldsymbol{\tilde{p}}$ is defined as a characteristic reference point on the body. When $\boldsymbol{p}_\perp$ is within the spherical region with radius $b$, then $\boldsymbol{\tilde{p}} = \boldsymbol{p}_\perp$. Outside the sphere, $\boldsymbol{\tilde{p}}$ is the projection of $\boldsymbol{p}_\perp$ on the sphere's surface.}
	\label{fig:spherical_region}
	\vspace{-5pt}
\end{figure}

The calculation of $\boldsymbol{p}_\perp$ using \eqref{eq:nu} is degenerate for pure translations, i.e. when $\Vert \boldsymbol{\omega} \Vert = 0$. Because of this, approaching a pure translation can result in $\Vert \boldsymbol{p}_\perp \Vert$ approaching infinity:
\begin{equation}
	\text{if}~~ \lim\limits_{\Vert \boldsymbol{\omega}\Vert \rightarrow 0}  \frac{\boldsymbol{\omega}\times\boldsymbol{v}}{\Vert \boldsymbol{\omega} \Vert} \neq \boldsymbol{0}~~, ~~\text{then}~~~ \lim\limits_{\Vert \boldsymbol{\omega}\Vert \rightarrow 0} \Vert \boldsymbol{p}_\perp \Vert = \infty.
\end{equation} 
To ensure that the calculated translational velocity of the rigid body and corresponding progress rate remain well-defined in these degenerate cases, a regularized translational velocity $\boldsymbol{\tilde{\nu}}$ is introduced, such that:
\begin{equation}
	\label{eq:reg_nu}
	\dot{s} = \sqrt{L^2\Vert \boldsymbol{\omega} \Vert^2 + \Vert \boldsymbol{\tilde{\nu}} \Vert^2}
	~~~\text{with}~~~
	\boldsymbol{\tilde{\nu}} = \boldsymbol{v} + \boldsymbol{\omega}\times\boldsymbol{\tilde{p}}~,
\end{equation}
where $\boldsymbol{\tilde{p}}$ is a regularized version of $\boldsymbol{p}_\perp$, defined as follows: 
\begin{equation}
	\label{eq:threshhold}
	\boldsymbol{\tilde{p}} = \left\{ 
	\begin{aligned}
		~~&\boldsymbol{p}_\perp &&\text{if~}   \Vert\boldsymbol{p}_\perp\Vert \leq b\\ \vspace{0pt}
		& \displaystyle b\frac{\boldsymbol{p}_\perp}{\Vert\boldsymbol{p}_\perp\Vert} &&\text{if~} \Vert\boldsymbol{p}_\perp\Vert > b\\
		&\boldsymbol{0} &&\text{if~} \Vert\boldsymbol{\omega}\Vert = 0.
	\end{aligned}
	\right.
\end{equation}
In other words, a sphere with radius $b$  is defined with the origin of the body frame as its center, so that when the position $\boldsymbol{p}_\perp$ is outside of the sphere, it will be projected onto the sphere's  surface. This is also visualized in Fig.~\ref{fig:spherical_region}.

The regularization in \eqref{eq:threshhold} has two important properties. First, the profile of $\boldsymbol{\tilde{\nu}}$ remains continuous over the threshold where $\Vert\boldsymbol{p}_\perp\Vert = b$, since:
\begin{equation}
	\lim\limits_{\Vert \boldsymbol{p}_\perp \Vert \rightarrow b^+} \boldsymbol{\tilde{p}} ~= \lim\limits_{\Vert \boldsymbol{p}_\perp \Vert \rightarrow b^-} \boldsymbol{\tilde{p}}~. 
\end{equation}
Second, the regularization is independent of the execution speed of the motion, since $\boldsymbol{p}_\perp$ is a geometric property.

In the Appendix, it is shown that choosing $b~\hspace{-3pt}\leq~\hspace{-4pt}L$ ensures that the triangle inequality \mbox{$\dot{s}_{1+2} \leq \dot{s}_1 + \dot{s}_2$} holds within the bi-invariant region $\Vert\boldsymbol{p}_\perp\Vert \leq b$, for the example discussed in Section \ref{sec:progress_rate}. As a result, we propose to choose $b=L$, since this results in the largest region of bi-invariance of the progress rate $\dot{s}$ without violating the triangle inequality. 

Remark that when $\Vert \boldsymbol{p}_\perp \Vert > b$, the regularized translational velocity $\boldsymbol{\tilde{\nu}}$ becomes dependent on the location of the body reference point and hence loses its \textit{bi-invariant} property.  This is not a big problem, since the object is dominantly translating in that case. Hence, the sensitivity of the velocity $\boldsymbol{\tilde{\nu}}$ to the location of the body reference point remains limited.


%

Algorithm~\ref{alg:xi} contains pseudocode to calculate the regularized progress rate $\dot{s}$ assuming $b=L$.


\vspace{-0pt}
\begin{algorithm}[h]
	\label{alg:progress}
	\caption{Calculation regularized progress rate $\dot{s}$}\label{alg:xi_dot}
	\label{alg:xi}
	\KwData{~\hspace{2pt}$\boldsymbol{\omega} \in \mathbb{R}^{3}$, $\boldsymbol{v} \in \mathbb{R}^{3}$, $L \in \mathbb{R}$}
	\KwResult{$\dot{s} \hspace{2pt}\in \mathbb{R}$, $\hspace{4pt}\tilde{\boldsymbol{p}} \in \mathbb{R}^{3}$, $\tilde{\boldsymbol{\nu}} \in \mathbb{R}^{3}$}
	\eIf{$\Vert \boldsymbol{\omega} \Vert = 0$}{
		$\boldsymbol{\tilde{p}} \gets \boldsymbol{0}$ ;
		$\boldsymbol{\tilde{\nu}} \gets \boldsymbol{v}$ ;
	}{$\boldsymbol{\tilde{p}} \gets (\boldsymbol{\omega} \times \boldsymbol{v})/(\boldsymbol{\omega}\cdot\boldsymbol{\omega})$ \;
		\lIf{$\Vert \boldsymbol{\tilde{p}} \Vert > L$}{
			$\boldsymbol{\tilde{p}} \gets L \boldsymbol{\tilde{p}}/\Vert \boldsymbol{\tilde{p}} \Vert$
		}
		$\boldsymbol{\tilde{\nu}} \gets \boldsymbol{v} + \boldsymbol{\omega} \times \boldsymbol{\tilde{p}}$ ;
	}
	$\dot{s} \gets \sqrt{L^2\boldsymbol{\omega}\cdot\boldsymbol{\omega} + \boldsymbol{\tilde{\nu}}\cdot\boldsymbol{\tilde{\nu}}}$ ; \vspace{-0pt}
\end{algorithm}
\vspace{-7pt}

\subsection{Reparameterization to geometric domain}
\label{sec:reparam}

The temporal rigid-body trajectory $T(t)$ can now be reparameterized to a geometric trajectory $T(s) = T(t(s))$ using the geometric progress rate $\dot{s}$ defined in \eqref{eq:reg_nu}. As a result, the execution speed will be separated from the geometric path of the trajectory. Such a geometric trajectory $T(s)$ is also referred to as a \textit{unit-speed} trajectory, since the geometric derivative $s' \left(=\frac{ds}{ds}\right)$ of the progress $s$ along this trajectory is equal to one:
\begin{equation} 
	\label{eq:norm1}
	s' = 1 = \sqrt{L^2 \left\Vert \boldsymbol{\omega}(s) \right\Vert^2 + \left\Vert \boldsymbol{\tilde{\nu}}(s) \right\Vert^2}.
\end{equation}



In practice, for discrete measurement data, this reparameterization is performed in three steps. Firstly, the temporal pose twist $\boldsymbol{\mathcal{t}}$ is calculated from the temporal pose trajectory $T(t)$ by numerical differentiation with the matrix logarithm operator \cite{lynch2017modern}. Secondly, the progress rate $\dot{s}$ is determined using Algorithm \ref{alg:progress}. The progress $s$ along the discrete trajectory is then found by a cumulative sum on $\dot{s}$. Thirdly, the trajectory $T$ is reparameterized from time $t$ to progress $s$ with fixed progress step $\Delta s$ using \textit{Screw Linear Interpolation}~\cite{kavan2008geometric}.

\subsection{Screw-based trajectory-shape descriptor}
\label{sec:descriptor}

This subsection introduces a screw-based trajectory-shape descriptor based on the rotational and translational velocities $\boldsymbol{\omega}(s)$ and $\boldsymbol{\tilde{\nu}}(s)$ along the reparameterized trajectory. 
These velocities can be calculated similarly to the calculation of $\boldsymbol{\omega}(t)$ and $\boldsymbol{\tilde{\nu}}(t)$, but now starting from the reparameterized trajectory $T(s)$ instead of $T(t)$. Afterwards, a normalization step is included to ensure that property \eqref{eq:norm1} holds. 

%

The proposed trajectory-shape descriptor at a given trajectory sample $i$ is of third order, and consists of the rotational and translational velocities at subsequent samples, centered at the sample point $i$, and stacked into a $3\times6$ matrix $S_i$: 
\begin{equation}
	\label{eq:shape_matrix}
	S_i = \begin{bmatrix} L\boldsymbol{\omega}_{i-1} & L\boldsymbol{\omega}_{i} & L\boldsymbol{\omega}_{i+1} & \boldsymbol{\tilde{\nu}}_{i-1} & \boldsymbol{\tilde{\nu}}_{i} & \boldsymbol{\tilde{\nu}}_{i+1} \end{bmatrix}.
\end{equation}


Since the velocities in \eqref{eq:shape_matrix} have their coordinates expressed with respect to some coordinate frame, a relative rotation alignment is needed before two local trajectory-shape descriptors can be compared. Similarly to the rotation alignment proposed in \cite{DSRF}, shape descriptor $S_2$ can be aligned with $S_1$ in three steps:
\begin{enumerate}
	\item Obtain the singular value decomposition of the relative $3\times3$ matrix $S_1 S_2^T = U \Sigma V$.
	\item Calculate the rotation  matrix $R = VU^T$, while ensuring that $R\in SO(3)$ by changing the sign of the third column of $U$ when \mbox{$\det(VU^T) = -1$}.
	\item Align $S_1$ with $S_2$ by left-multiplying $S_1$ with $R$.  
\end{enumerate} 

The difference in local trajectory-shape $\Delta_1^2 S$ of two descriptors $S_1$ and $S_2$ is then defined as the Frobenius norm of the difference between the descriptors after alignment: 
\begin{equation}
	\label{eq:shape_metric}
	\Delta_1^2 S = \left\Vert RS_1 - S_2 \right\Vert_F ~.
\end{equation}
The difference $\Delta_1^2 S$ is a value with units [m], measuring the difference in local change of the trajectory shape.



\section{APPLICATION TO SEGMENTATION}
\label{sec:segmentation}

This section applies the proposed screw-based progress rate $\dot{s}$ and trajectory-shape descriptor $S$ to a trajectory segmentation approach. Envisioned towards incremental learning applications, a self-supervised segmentation approach based on incremental clustering was devised.
This segmentation approach consists of two phases: an offline learning phase where a library of trajectory-shape primitives is learned, and a trajectory segmentation phase. 
%



\textbf{Offline template learning}:
Trajectory-shape primitives are learned by an incremental clustering approach, where the mean of each cluster represents a learned primitive. 
The clustering approach consists of four steps:

1) \textit{Initialization:} Use the first sample point $S_1$ to generate the first cluster with mean $S_1$ and a chosen initial guess for its standard deviation $\sigma_0$. 

2) \mbox{\textit{Cluster growing:}} Given a sample point $i$ with corresponding trajectory-shape descriptor $S_i$, calculate the difference $\Delta S$ between the descriptor $S_i$ and the mean $\bar{S}$ of every learned cluster. Afterwards, add the sample point to the cluster with the smallest difference in trajectory-shape, if this difference is smaller than three times its standard deviation $\sigma$, and update the mean $\bar{S}$ and standard deviation $\sigma$ of the cluster accordingly. If no such cluster exists, create a new cluster with mean $S_i$ and initial standard deviation $\sigma_0$.

3) \mbox{\textit{Cluster parameter update:}} After all the available data is clustered, update the value for $\sigma_0$ based on the mean of the standard deviations of all the learned clusters~$\bar{\sigma}$:  \mbox{$~\sigma_{0,next} = \bar{\sigma}~+~ \hat{\sigma}$}, with $\hat{\sigma}$ a tuning parameter related to the process noise of the clusters. Iterate steps 1 to 3 until $\sigma_0$ converges to a steady value.

4) \mbox{\textit{Outlier removal:}} After convergence, remove sparse clusters that do not represent at least $\beta \%$ of the data, with $\beta$ being a chosen value.


\textbf{Trajectory segmentation phase}:
Each sample point gets associated with a learned primitive (using a \mbox{1-Nearest-Neighbor} classifier) as long as the difference in trajectory-shape $\Delta S$ is smaller than $3\sigma$ of the respective cluster. Otherwise, the sample point is labeled as `\mbox{non-classified}'. Afterwards, segments are formed by grouping consecutive sample points along the trajectory that were associated with the same trajectory-shape primitive. 


\section{EXPERIMENTS}
\label{sec:experiments}


A key property of the proposed trajectory representation is its invariance to time and the choice of body reference point. The main aim of the experiments is to demonstrate the advantages of these invariant properties for trajectory segmentation. This is done both in simulation and using real recordings of human-demonstrated pouring motions.

\begin{figure}[t]
	\centering
	\vspace{5pt}
	\includegraphics[width=0.65\linewidth,trim = 0 0 1580 0, clip]{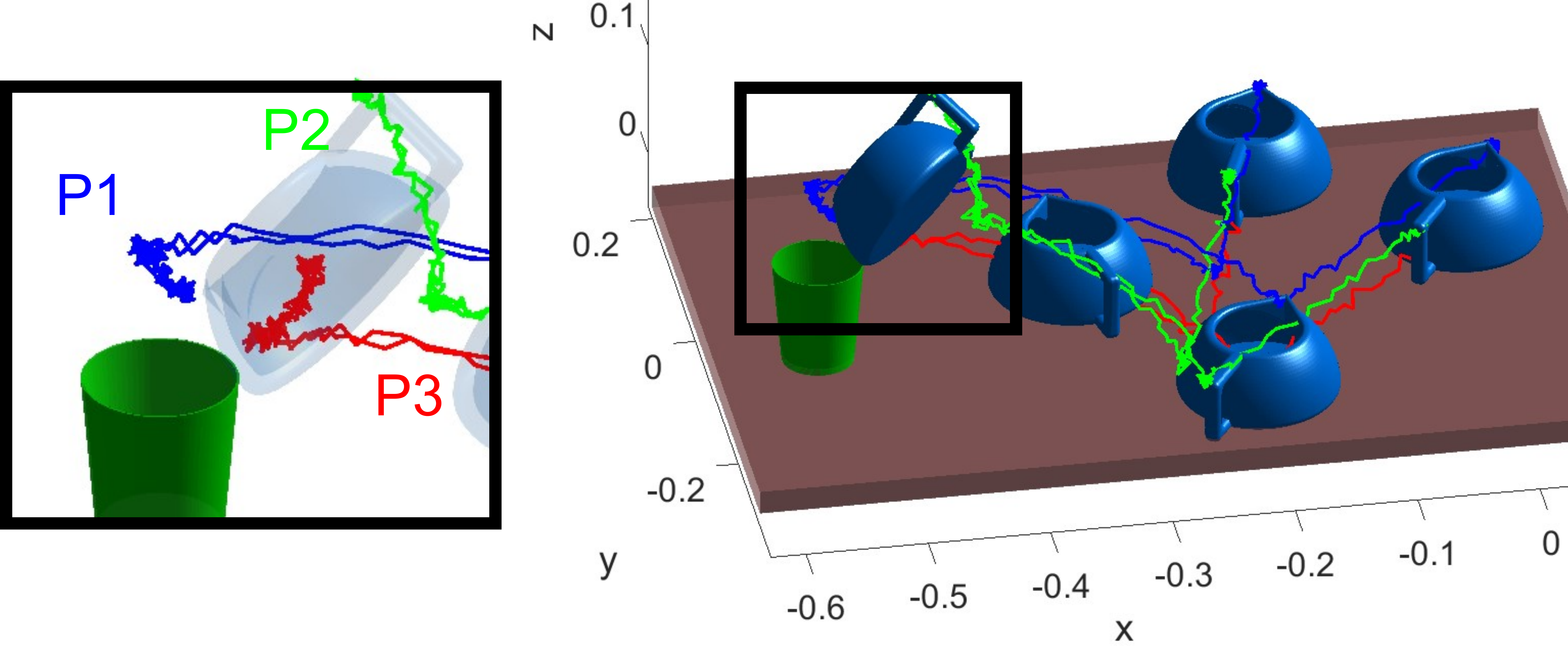}%
	\caption{Visualization of three rigid-body trajectories (red, green, and blue) representing simulated pouring motions performed with a kettle. Different body reference points (\color{black} \color{blue} P1\color{black}, \color{green} P2 \color{black} and \color{red} P3\color{black}) were considered.}
	\label{fig:simulated_pouring_motion}
\end{figure}

The simulated trajectories represent temporal rigid-body trajectories $T(t)$ of pouring motions performed with two types of objects: a teakettle and a bottle. The strategy of the pouring motion consists of a sequence of six intuitive sub-motions, also visualized in Fig.~\ref{fig:simulated_pouring_motion}:
\begin{itemize}
	\item{\makebox[1cm][l]{\textit{slide+}} : the object is slid across the table from its initial position to a fixed location on the table}
	\item{\makebox[1cm][l]{\textit{lift+}} : the object is lifted and reoriented so that the spout is directed towards the glass}
	\item{\makebox[1cm][l]{\textit{tilt+}} : the object is tilted to pour the liquid}
	\item{\makebox[1cm][l]{\textit{tilt-}} :  the object's tilt is undone}
	\item{\makebox[1cm][l]{\textit{lift-}} : the object is placed back on the table}
	\item{\makebox[1cm][l]{\textit{slide-}} : the object is slid back}
\end{itemize}
We aim to robustly segment the pouring motion into these sub-motions. To simulate sensor noise, white noise with a standard deviation of 2$\degree$ and 1 mm were added to the object's orientation and reference point's location, respectively. 

To study robustness to changes in the body reference point, each of the three trajectories was simulated with a different reference point: the first was near the spout (\color{blue}{P1}\color{black}), the second near the handle (\color{green}{P2}\color{black}), and the third near the center of mass (\color{red}{P3}\color{black}). Fig.~\ref{fig:simulated_pouring_motion} on the left illustrates that, when rotation of the object is involved, the different reference points result in significantly different point trajectories. 

The generalization capability is tested by applying the primitives that were learned for the kettle motion to another object. The object was chosen to be a bottle of which the reference point was chosen in the center of its opening. 

The proposed approach is compared to other methods based on the literature, shown in Table~\ref{tab:hyperparameters}.
\textit{Method $A$} does not transform the object's trajectory to a geometric domain.
\textit{Methods $A$ and $B$} neglect information on the object's orientation by choosing $L = 0$.
\textit{Methods $B$ and $C$} use the arclength of the reference point's trajectory as the geometric progress parameter, such as in \cite{DSRF}.
\textit{Method $D$} uses the traveled angle of the moving object as the geometric progress parameter, such as in \cite{Roth2005}.
\textit{Method $E$} uses a combination of the rotational velocity and linear velocity of the reference point on the body as the progress rate, such as in \cite{SE3metricsPark}.
\textit{Method $F$} uses a screw-based progress rate without the proposed regularization, such as in \cite{Joris2010}.
\textit{Method $G$} uses the proposed screw-based geometric progress rate. 



\begin{table}[t]
	\centering
	\caption{Summary of methods and corresponding tuning parameters}
	\label{tab:hyperparameters}
	\resizebox{\linewidth}{!}{
		\begin{tabular}{clccccc}
			method & progress-type & $\dot{s}$ &  $L$ [cm] & $\Delta s$ & $\hat{\sigma}$ & $\beta$ \\ \midrule
			A & time & $1$ & 0 & 0.1s & 10cm/s & 2\% \\
			B & arclength  & $\Vert \boldsymbol{v} \Vert $ & 0 & 2cm & 2cm & 5\% \\
			C & arclength & $\Vert \boldsymbol{v} \Vert $ & 30 & 2cm & 10cm & 5\% \\
			D & angle  & $\Vert \boldsymbol{\omega} \Vert $ & 30 & 3\degree & 10cm & 5\% \\
			E  & combined  & $\sqrt{L^2\Vert \boldsymbol{\omega} \Vert^2 + \Vert \boldsymbol{v} \Vert^2} $ & 30 & 2cm & 10cm & 5\% \\
			F & screw-based  & $L \Vert \boldsymbol{\omega} \Vert + \Vert \boldsymbol{\nu} \Vert $ & 30 & 2cm & 10cm & 5\% \\
			\textbf{G} & \textbf{screw-based}  & $\sqrt{L^2 \Vert \boldsymbol{\omega} \Vert^2 + \Vert \boldsymbol{\tilde{\nu}} \Vert^2} $ & \textbf{30} & \textbf{2cm} & \textbf{10cm} & \textbf{5\%} 
		\end{tabular}
	}\vspace{-5pt}
\end{table}

\begin{figure}[t]
	\centering
	\subfloat[]{\includegraphics[width=0.25\linewidth]{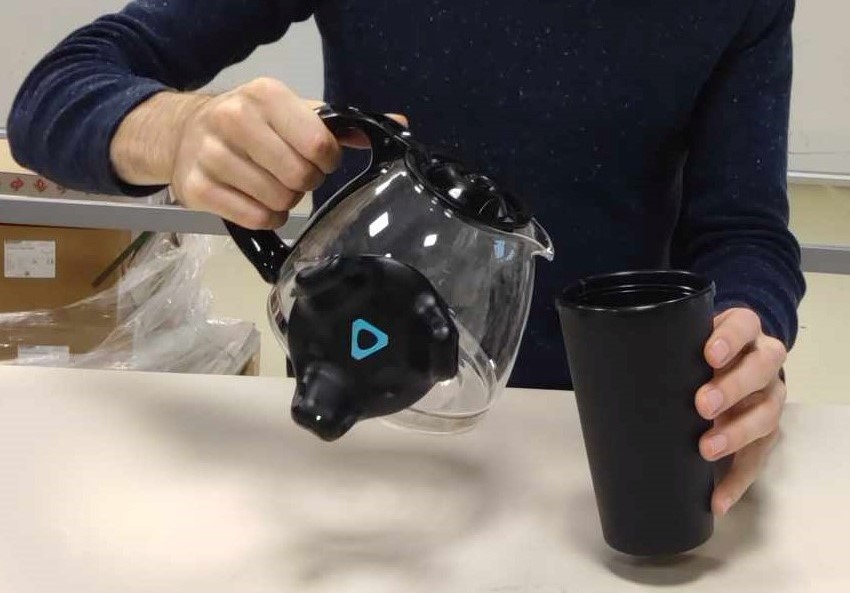}%
		\label{fig:foto_real_recording}}
	\hfil
	\subfloat[]{\includegraphics[width=0.3\linewidth,trim = 0 0 0 0, clip]{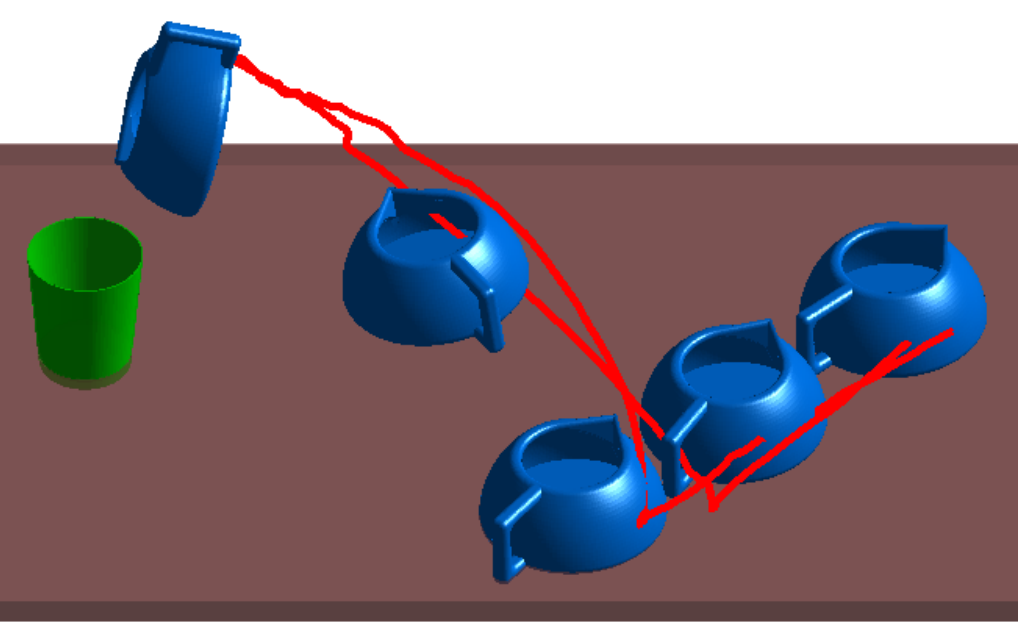}%
		\label{fig:trajectory_real_recording}}
	\caption{(a) Human demonstration of a pouring motion using a teakettle to which an HTC VIVE tracker is attached. (b) Visualization of the first trial within a batch of six trials in the same simulation environment as Fig.~\ref{fig:simulated_pouring_motion}. 
	}
	\label{fig:test}
	\vspace{-0pt}
\end{figure}

To validate that the proposed method works in practice, it was tested on recordings of real pouring motions. These motions were recorded using an HTC VIVE motion capture system, consisting of a tracker attached to the kettle (see
Fig.~\ref{fig:foto_real_recording}), and two base stations. The VIVE system recorded the pose trajectories with a frequency of 60 Hz and an accuracy in the order of a few mm and a few degrees. To introduce contextual variations in the measurements, the tracker was physically attached to the kettle at two different locations, one at the side of the kettle (P1) and one near the top of the handle (P2). For each tracker location, three trials were recorded. Fig.~\ref{fig:trajectory_real_recording} depicts the kettle's pose trajectory for the first trial with the tracker attached to the side.

\subsection{Data processing}

The simulated data and the recorded data were processed in the same way. The rigid-body trajectories were first preprocessed using a Kalman smoother with a constant acceleration model to deal with the effects of the measurement noise. For the methods $B$ to $G$, the trajectories were first reparameterized as in Section~\ref{sec:progress_rate} according to the chosen definition of the progress rate $\dot{s}$. Then, the local shape descriptor was calculated as in Section~\ref{sec:descriptor}. Finally, the same segmentation algorithm (Section~\ref{sec:segmentation}) is applied for all approaches.
For reproducability reasons, all software used in the experiments is made publicly available \cite{software}.

The values of the tuning parameters are reported in Table~\ref{tab:hyperparameters}. A good choice for $L$ depends on numerous factors, including the scale of the motion and the scale of the moving object. 
Given the scale of the objects and motions of interest, a value of $L=30$ cm seemed reasonable. The other parameters in Table~\ref{tab:hyperparameters} were manually tuned. For method $A$, the parameters $\Delta s$ and $\hat{\sigma}$ have different units compared to the ones of the other methods since method $A$ does not transform the trajectories to a geometric domain before segmentation. 

\subsection{Results}

\begin{figure}[t]
	\centering
	\vspace{5pt}
	\includegraphics[width=0.95\linewidth]{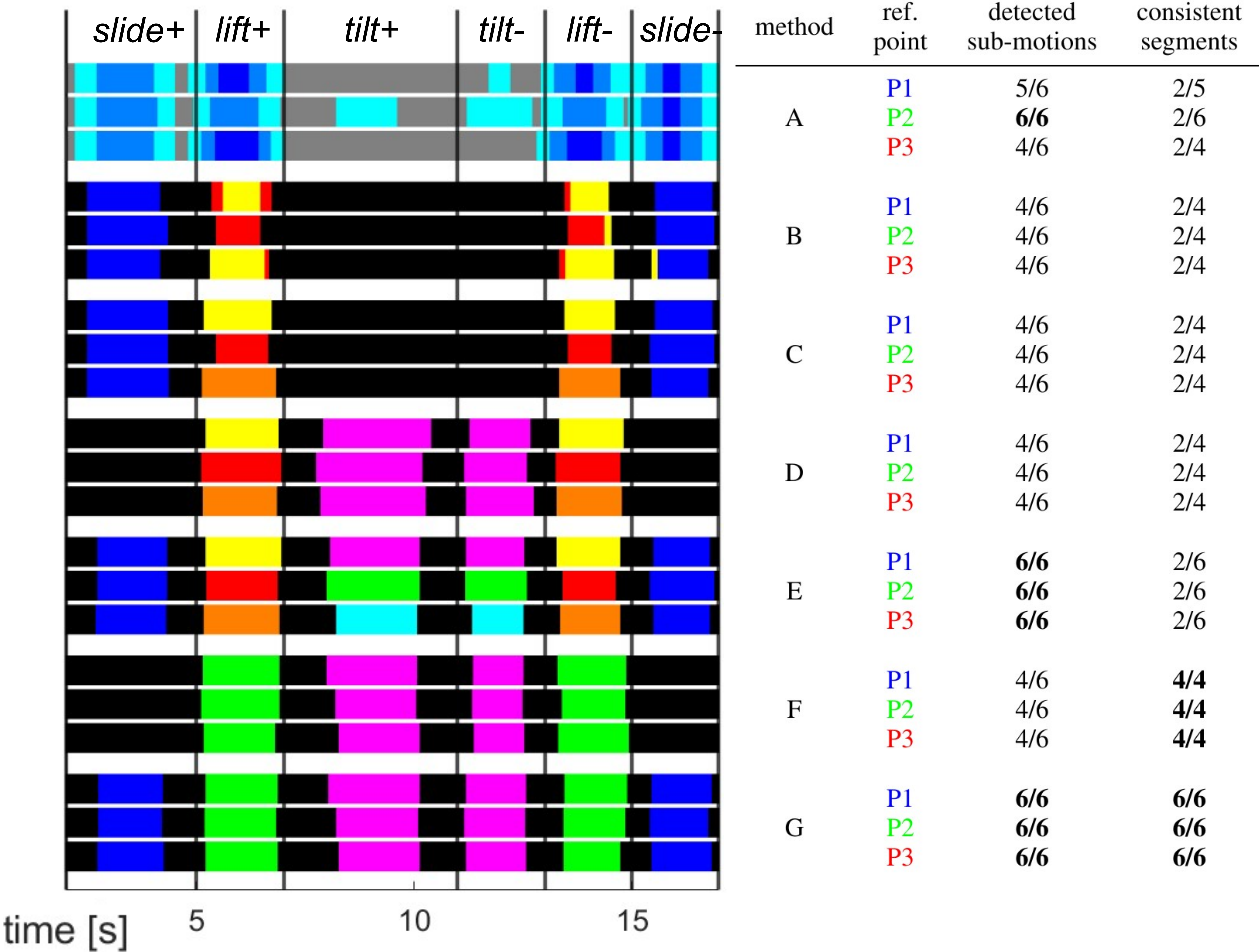}%
	\caption{Comparison of segmentation results between methods \textit{$A$-$G$} for the simulated pouring motion data. Segments associated with different primitives are indicated with different colors. The black regions contain non-classified samples.}
	\label{fig:evaluation}
	\vspace{-3pt}
\end{figure}

\textbf{Simulated data}:
Fig.~\ref{fig:evaluation} visualizes the segmentation results of all methods for the simulated data. The ground-truth segmentation points are indicated by the vertical lines.
To compare between methods, the segmented trajectories were transformed back to the time domain by re-applying the motion profile $s(t)$ that was extracted from the trajectories. This was done by inverting the reparameterization procedure of Section~\ref{sec:reparam}. The segmentation results are also evaluated quantitatively by reporting the number of \textit{detected sub-motions} and \textit{consistent segments}. A sub-motion was considered `detected' when a corresponding segment was formed. A sub-motion was considered `consistently segmented' when the corresponding segments were associated to the same trajectory-shape primitive across the three trials.

The results of the simulation are interpreted as follows. Method $A$ generated a relatively high number of segments. The segmentation is mainly based on differences in magnitude of the reference point's velocity. The gray segments represent regions of standstill. The light and dark blue segments represent segments of low and high magnitude in velocity, respectively. Methods $B$ to $G$ generated segments based on differences in shape of the rigid-body trajectory.


For methods \mbox{$A$-$E$}, the learning of the primitives and segmentation of the trajectories was dependent on the location of the reference point on the object. Furthermore, methods $A$-$C$ could not deal well with pure rotations of the object. Method $A$ classified these segments either as stationary or as segments with low magnitude in velocity. Methods $B$ and $C$ treated these segments as outliers. The reason for this is that during these pure rotations, the traveled arclength of the reference point remained relatively small, resulting in a small number of geometric sample points representing these pure rotations. Hence, for pure rotations, sparse clusters were created, which were seen as outliers. Following a similar reasoning, method $D$ could not deal with pure translations.


Method $F$ performed the segmentation in a reference-point invariant way, but could not deal well with pure translations, since $\boldsymbol{p}_\perp$ is degenerate in this case.

\begin{figure}[t]
	\centering
	\includegraphics[width=0.9\linewidth,trim = 0 0 0 0, clip]{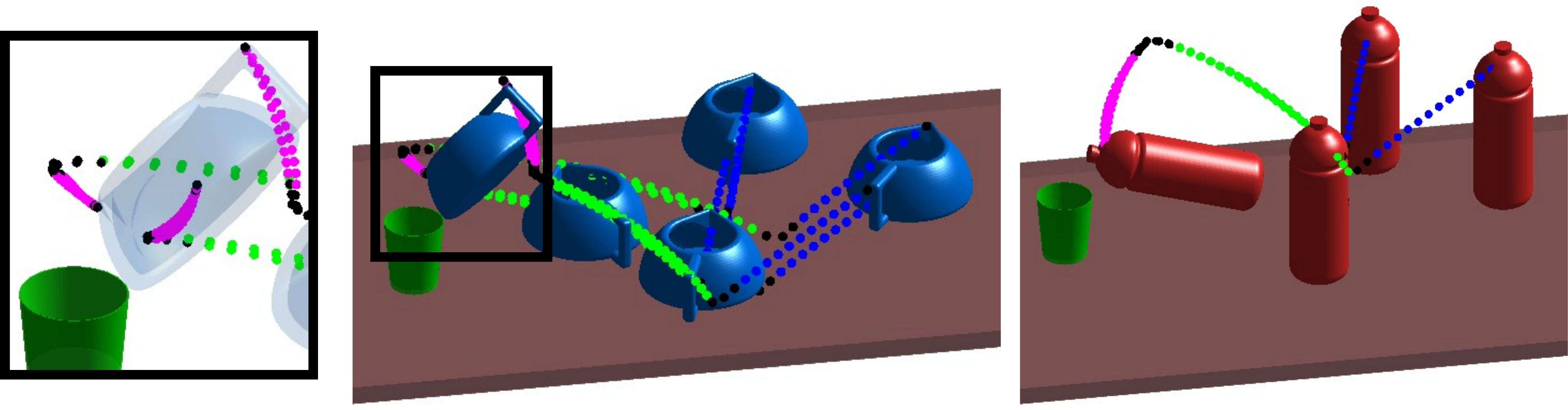}%
	\caption{Visualization of the segmented trajectories using the proposed method (method $G$).}
\label{fig:segments}
\vspace{-8pt}
\end{figure}

The proposed method $G$ dealt well with pure translations and performed the segmentation in a reference-point invariant way. The trajectories of the kettle were consistently segmented into six segments, corresponding to the six intuitive sub-motions. More in detail, all sub-motions were detected (6/6) and all sub-motions were consistently segmented (6/6).
Fig.~\ref{fig:segments} visualizes the segmented trajectories. 


The proposed approach also succeeded to segment the simulated pouring motion performed with a bottle (with significantly different location of the reference point) using the trajectory-shape primitives learned from the trajectories of the kettle. This illustrates the capability of the approach to generalize to different objects with different geometries.

\begin{figure}[t]
	\centering
	\vspace{5pt}
	\includegraphics[width=\linewidth]{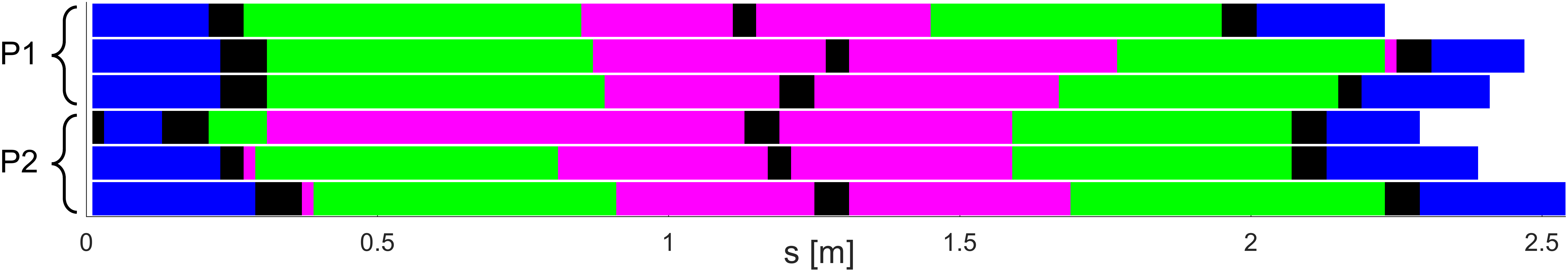}%
	\caption{Segmentation results of proposed method $G$ for the real recorded pouring motion data. 
	P1 and P2 correspond to trials with a different location of the motion tracker on the kettle.
	}
	\label{fig:eval_real recording}
	\vspace{-0pt}
\end{figure}

\textbf{Real data}:
Fig.~\ref{fig:eval_real recording} visualizes the segmentation results of the proposed method $G$ for the real recorded pouring motions. To illustrate the generated segments in the geometric domain, the extra transformation back to the time domain was not performed. The same values for the tuning parameters as reported in Table~\ref{tab:hyperparameters} were used for this experiment.   
%

Three primitives were learned from the real data. The corresponding values of the mean trajectory-shape descriptor $\bar{S}$ of the three clusters are reported in Table~\ref{tab:primitives_real}.
The learned primitives represent 1D motions, since for each primitive, the three columns are almost identical. The first primitive represents a 1D translation (slide). The second primitive represents a rotation with a non-zero pitch (lift). The third primitive represents a pure rotation (tilt). The segmentation approach created segments conform the six intuitive sub-motions, apart from some short segments near transition regions. 
These short segments can be avoided by implementing a postprocessing step or a more advanced segmentation algorithm from the literature, which is part of future work.


\begin{table}[t]
	\centering
	\caption{Cluster means learned from the real pouring motion data.}
	\label{tab:primitives_real}
	\resizebox{1\linewidth}{!}{
		\begin{tabular}{c | ccc | ccc | ccc}
			\vspace{0pt} 
			& \multicolumn{3}{c|}{\large\color{blue}$\bar{S}_1$} & \multicolumn{3}{c|}{\large\color{green}$\bar{S}_2$} & \multicolumn{3}{c}{\large\color{magenta}$\bar{S}_3$} \\ 
			& $s-\Delta s$ & $s$ & $s+\Delta s$ & $s-\Delta s$ & $s$ & $s+\Delta s$ & $s-\Delta s$ & $s$ & $s+\Delta s$ \\ \midrule
			$L\omega_x$ & -0.02 & -0.02 & -0.02 & \textbf{0.97} & \textbf{0.97} & \textbf{0.97}\ & \textbf{1.00} & \textbf{1.00} & \textbf{1.00}   \\
			$L\omega_y$ & -0.01 & -0.01 & -0.01 & -0.03 & -0.01 & -0.02 & 0.01 & -0.00 & -0.01  \\
			$L\omega_z$ & 0.12 & 0.11 & 0.11 & 0.02 & 0.01 & 0.01 & 0.03 & -0.00 & -0.03  \\
			$\nu_x$ & \textbf{0.99} & \textbf{0.99} & \textbf{0.99} & \textbf{0.26} & \textbf{0.26} & \textbf{0.26} & 0.06 & 0.06 & 0.06  \\
			$\nu_y$ & 0.01 & 0.01 & 0.01 & -0.00 & 0.01 & 0.01 & 0.00 & 0.00 & -0.00  \\
			$\nu_z$ & -0.11 & -0.11 & -0.11 & -0.01 & -0.02 & -0.01 & 0.00 & 0.00 & -0.00  \\
		\end{tabular}
	}
	\vspace{-5pt}
\end{table}

\section{DISCUSSION AND CONCLUSION}
\label{sec:discussion}


The objective of this work was to enhance template-based trajectory segmentation approaches by incorporating invariance. Time-invariance was achieved by reparameterizing the trajectory using a novel geometric progress parameter. By considering the translation along the screw axis, the progress parameter was made invariant to the choice of body reference point. Based on the reparameterized trajectory, a screw-based trajectory-shape descriptor was proposed to characterize the local geometry of the trajectory. 


For the devised self-supervised segmentation scheme, the results showed a more robust detection of consecutive sub-motions with distinct features and a more consistent segmentation thanks to the invariant properties of the screw-based progress parameter and the trajectory-shape descriptor.



The proposed approach also has a more practical advantage. Due to the invariance, the formation of the segments becomes invariant to changes in sensor setup (i.e. changes in the location and angle of the camera, changes in the location of markers or trackers on the object, etc.). Therefore, sensor calibration efforts can be reduced.






%
Future work is to examine the benefits of the invariant segmentation approach for other types of object manipulation tasks and to verify the extent to which other segmentation methods may benefit from the invariant approach.



%

\section*{APPENDIX}
\label{app}

This appendix shows that the triangle inequality property $\dot{s}_{1+2} \leq \dot{s}_1 + \dot{s}_2$ in Section \ref{sec:regul_progress} is always satisfied for the regularized progress rate $\dot{s}$ if $b$ equals $L$ in \eqref{eq:threshhold}. 

Consider again the case of a couple of rotations generating a translation as explained in Section \ref{sec:regul_progress}. Additionally consider that the rotation axes of the couple are symmetrically positioned w.r.t. the reference point, such that $a = \Vert \boldsymbol{p}_\perp \Vert$. Equation \eqref{eq:example_couple} then remains valid under the proposed regularization action when $\Vert \boldsymbol{p}_\perp \Vert \leq b$. From \eqref{eq:example_couple}, it was derived that the triangle inequality holds when $ a \leq L$. Hence, given that $a = \Vert \boldsymbol{p}_\perp \Vert$ and $\Vert \boldsymbol{p}_\perp \Vert \leq b$, then $ a \leq L$ is always satisfied when $b \leq L$.

\newpage

\bibliographystyle{IEEEtran}
\bibliography{bibfile}

\end{document}